\documentclass{article}
\usepackage{enumitem}
\usepackage{wrapfig}    
\usepackage{adjustbox}  
\usepackage{multirow}
\usepackage{makecell}
\usepackage{natbib}
\usepackage{subcaption}
\usepackage{graphicx}
\usepackage{tabularx}
\usepackage{caption}

\usepackage[preprint]{neurips_2025}

\usepackage[utf8]{inputenc} 
\usepackage[T1]{fontenc}    
\usepackage{hyperref}       
\usepackage{url}            
\usepackage{booktabs}       
\usepackage{amsfonts}       
\usepackage{nicefrac}       
\usepackage{microtype}      
\usepackage{xcolor}         
\usepackage{graphicx}

\usepackage{arydshln} 
\usepackage{longtable}  
\usepackage{geometry}   

\title{SurgBench: A Unified Large-Scale Benchmark for Surgical Video Analysis}

\author{%
  Jianhui Wei\textsuperscript{1,4,\thanks{Co-first author.}} \quad
  Zikai Xiao\textsuperscript{1,4,\footnotemark[1]} \quad
  Danyu Sun\textsuperscript{1} \quad
  Luqi Gong\textsuperscript{2} 
  \\
  \textbf{Zongxin Yang}\textsuperscript{3} \quad
  \textbf{Zuozhu Liu}\textsuperscript{1,4,\thanks{Co-corresponding author.}} \quad
  \textbf{Jian Wu}\textsuperscript{1,4,\footnotemark[2]}
  \\[0.5ex]
  \textsuperscript{1}Zhejiang University \quad
  \textsuperscript{2}Zhejiang Lab \quad
  \textsuperscript{3}Harvard University
  \\
  \textsuperscript{4}Zhejiang Key Laboratory of Medical Imaging Artificial Intelligence
  \\[0.3ex]
  $^{*}$\texttt{jianhui1.24@intl.zju.edu.cn} \quad
  $^{\dagger}$\texttt{zuozhuliu@intl.zju.edu.cn}
}

\begin{document}

\maketitle

\begin{abstract}

Surgical video understanding is pivotal for enabling automated intraoperative decision-making, skill assessment, and postoperative quality improvement. However, progress in developing surgical video foundation models (FMs) remains hindered by the scarcity of large-scale, diverse datasets for pretraining and systematic evaluation. In this paper, we introduce \textbf{SurgBench}, a unified surgical video benchmarking framework comprising a pretraining dataset, \textbf{SurgBench-P}, and an evaluation benchmark, \textbf{SurgBench-E}. SurgBench offers extensive coverage of diverse surgical scenarios, with SurgBench-P encompassing 53 million frames across 22 surgical procedures and 11 specialties, and SurgBench-E providing robust evaluation across six categories (phase classification, camera motion, tool recognition, disease diagnosis, action classification, and organ detection) spanning 72 fine-grained tasks. Extensive experiments reveal that existing video FMs struggle to generalize across varied surgical video analysis tasks, whereas pretraining on SurgBench-P yields substantial performance improvements and superior cross-domain generalization to unseen procedures and modalities. Our dataset and code are available upon request.

\end{abstract}

\section{Introduction}
Surgical video analysis is rapidly becoming a cornerstone for advancing modern surgical care, offering a window into the intricacies of operative procedures \citep{green2019utilization}. The ability to systematically interpret these videos can significantly impact intraoperative decision-making, automate surgical skill evaluation, and inform postoperative quality improvement initiatives \citep{loftus2020decision, prebay2019video, grenda2016using, levin2019automated, gruter2023video, dimick2015surgical}.

In response to the clinical opportunities, video foundation models (VFMs) have emerged as a powerful framework for surgical video analysis. By leveraging large-scale, diverse datasets, VFMs enable efficient and accurate video modeling, enhancing the potential for surgical video analysis and improving surgical practice \cite{wang2024internvideo2, zhao2024videoprism, li2023unmasked, madan2024foundation}.
VFMs have demonstrated success in a variety of downstream tasks, benefiting from pre-training on massive datasets that span millions of video clips and several terabytes of data \citep{wang2024internvideo2, zhang2025videollama, tong2022videomae, wang2023videomae}. These datasets encompass diverse video categories, such as YouTube videos, movie trailers, and surveillance footage, enabling models to develop generalizable representations across domains \citep{li2023unmasked, madan2024foundation}.

Despite the proven efficacy of VFMs in general video analysis, their application to surgical video analysis remains in its infancy, primarily due to \textbf{ the limited diversity of disease types, surgical procedures, and specialties, and insufficient task comprehensiveness in existing pretraining and evaluation datasets}. Disease diversity is constrained, as most datasets focus on specific conditions, such as colorectal cancer~\citep{misawa2021development}, limiting model generalization across diverse anatomical variations, comorbidities, and rare diseases \citep{bar2020impact}. The coverage in surgical procedures or specialties is narrow, with datasets that predominantly feature specific specialties and minimally invasive procedures, such as laparoscopic cholecystectomy\citet{wang2022autolaparo, twinanda2016endonet} or robotic prostatectomy \citet{ahmidi2017dataset}, while under-representing open, hybrid, or less standardized surgical techniques.
Insufficient task comprehensiveness is apparent in the tendency to emphasize isolated analytical tasks, such as phase classification for a specific procedure\citep{goodman2024analyzing, fujii2024egosurgery} or instrument recognition \citep{ma2021ldpolypvideo}, while neglecting integrated clinical workflows that span multiple temporal and semantic dimensions. 

\begin{table}[t]
\centering
\resizebox{\linewidth}{!}{
\begin{tabular}{ccccc}
\toprule
\multirow{2}{*}{\textbf{Datasets}} & \multicolumn{3}{c}{\textbf{Pretraining Data}} & \multirow{2}{*}{\textbf{Evaluation Data}} \\
\cline{2-4}
& \# Surgical specialties& \# Surgical procedures & \# Frames & \\
\midrule
Endo-FM (\cite{wang2023foundation}) & 3 & 3 & 5M & \textcolor{red}{$\times$}  \\
GSViT (\cite{schmidgall2024general}) & 3 & 28 &  70M & \textcolor{red}{$\times$}  \\
Surg-3M (\cite{che2025surg}) & 5 & 35 & 3M & \textcolor{red}{$\times$} \\
\midrule
SurgBench (Ours) & 11 & 22 & 53M & \textcolor{green}{\checkmark} (72 Tasks)\\
\bottomrule
\end{tabular}
}
\vspace{5pt}
\caption{Comparison of datasets used for surgical video pretraining and evaluation. Our benchmark achieves advantages in the coverage of surgical specialties and surgical procedures, while also providing comprehensive evaluation standards for downstream tasks.}

\end{table}

In this paper, we introduce \textbf{SurgBench}, a unified, large-scale surgical video benchmarking framework consisting of a pretraining dataset, \textbf{SurgBench-P}, and an evaluation benchmark, \textbf{SurgBench-E}.
The pre-training subset \textbf{SurgBench-P} comprises 53 million frames from 16 distinct sources (Figure~\ref{fig:construction pipeline}), encompassing 4 primary surgical modalities: laparoscopic, endoscopic, robotic, and open surgery. It spans 22 diverse surgical procedures (e.g., appendectomy, cholecystectomy, colectomy, etc) across 11 medical specialties (see Table~\ref{tab:surgeries}), directly mitigating procedural and disease-type homogeneities. We further leverage uniform spatiotemporal protocols (e.g., length, frame rate, encoding), aligned with general video understanding benchmarks like Kinetics, across all video sources. This standardization ensures pre-training consistency and supports effective knowledge transfer from general-domain VFMs. Complementing the pre-training corpus, \textbf{SurgBench-E} serves as an integrated fine-tuning and hierarchical evaluation framework specifically designed to enrish task diversity and rigorously assess clinical utility. SurgBench-E features 23,004 surgical video clips with granular annotations for 6 distinct surgical understanding task categories, 10 sub-categories, and further delineated into 72 tasks, offering a structured means for comprehensive model evaluation, as detailed in Table~\ref{tab:task_codes}. 

Initial empirical validation underscores the efficacy of this approach: SurgMAE, pre-trained on SurgBench, yields performance gains of 7.9\% on downstream tasks relative to models trained on the natural video dataset Kinetics.
Our key contributions are as follows:

\begin{itemize}
    \item We introduce \textbf{SurgBench-P}, a large-scale and standardized pretraining dataset (53M frames, 22 procedures, 11 specialties, 4 modalities), designed to address procedural and disease-type homogeneity in surgical video analysis. 
    \item We construct \textbf{SurgBench-E}, a comprehensive evaluation benchmark encompassing 72 fine-grained tasks across 6 categories, mitigating task fragmentation and enabling systematic evaluation of clinical utility.
    \item We validate a self-supervised pretraining pipeline using VideoMAE for superior performance, demonstrating the potential of \textbf{SurgBench} to advance research and education in surgery.
\end{itemize}

\section{Related Work}
\label{sec:related_work}
\textbf{Surgical Video Analysis.} Early research in surgical video analytics focused on single diseases or a limited range of tasks, such as surgical phase recognition (\cite{yu2018learning}), instrument detection (\cite{yu2018learning}), and polyp detection (\cite{ma2021ldpolypvideo}). These approaches typically rely on small, single-center datasets, lacking diversity, which results in task- and scenario-dependent model training with limited generalization capabilities. For instance, studies have been conducted on endoscopic surgeries targeting specific conditions (\cite{nwoye2021rendezvous}), laparoscopic hysterectomy procedures (\cite{wang2022autolaparo}), colonoscopic videos for polyp detection (\cite{mesejo2016computer}, \cite{ma2021ldpolypvideo}), and capsule endoscopy using PillCAM data (\cite{Smedsrud2021}). Additionally, due to the high cost of annotation, many of these efforts are heavily supervised, making it challenging to scale to a broader range of clinical environments or multi-task learning scenarios.

\textbf{Surgical Foundation Models.} With increasing availability of large-scale surgical video data, researchers have developed Surgical Foundation Models \citep{wang2023foundation, schmidgall2024general}, often leveraging self-supervised learning on diverse datasets for generalizable representations. These models demonstrate superior robustness and cross-domain transferability. More recently, the integration of vision-language and embodied AI approaches \citep{li2024llava, wang2025endochat, li2025tacoenhancingmultimodalincontext, li2025m2ivefficientfinegrainedmultimodal, li2025camaenhancingmultimodalincontext, bi2024visual, bi2025prism, zeng2024mitigating, guo2025each, ma2024event, zhou2025dual, zhou2025ssfold} has further enhanced semantic understanding and interpretability for complex surgical tasks.

\section{Dataset Construction Pipeline}
\label{sec:sec3}
\begin{figure}[htbp]
  \centering
  \includegraphics[width=1.0\textwidth]{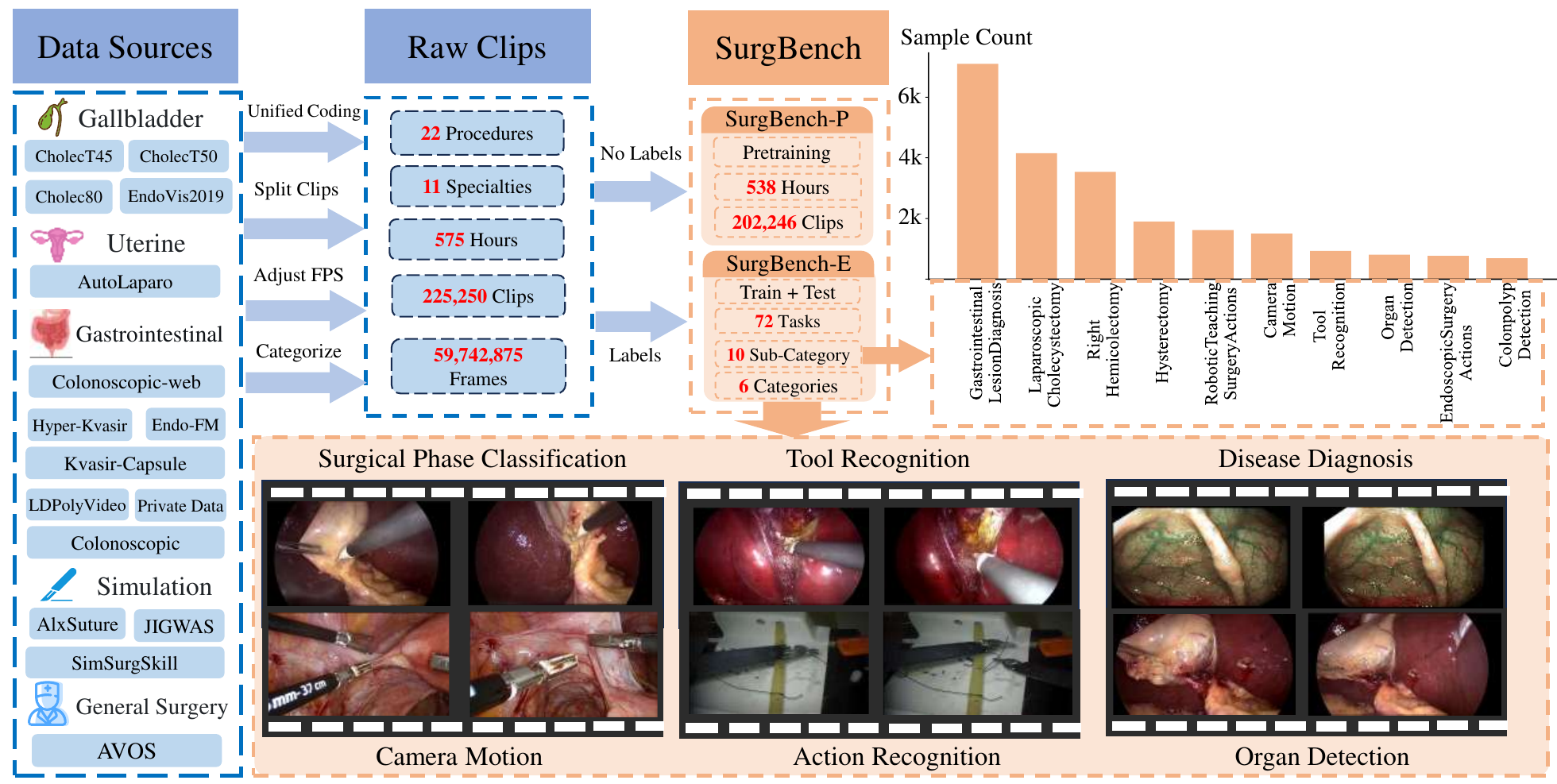}
    \caption{Data construction pipeline of SurgBench. We collect 16 datasets (including private datasets), standardized all video formats to accelerate training, and organized them into SurgBench-P (unlabeled data) and SurgBench-E (labeled data), containing 60 million frames in total. SurgBench-P covers 11 clinical specialties and 22 surgery procedures, while SurgBench-E encompasses 6 categories and 10 sub-categories, totaling 72 tasks. Clip examples are given at the bottom of the figure. The distribution of 10 sub-categories is long-tailed, as shown in the upper right of the figure.}
  \label{fig:construction pipeline} 
\end{figure}
\label{sec:dataset}
In this section, we introduce the construction pipeline of how we formulate SurgBench, which encompasses SurgBench-P for pretraining models and SurgBench-E for fine-tuning and evaluating models. The whole construction pipeline is shown in Figure \ref{fig:construction pipeline}.
\subsection{Dataset Source}

\begin{table*}[!ht]
\centering
\resizebox{\linewidth}{!}{
\scriptsize
\begin{tabular}{@{}p{0.5cm}p{2.5cm}p{2cm}p{2.6cm}p{2.6cm}p{1.4cm}p{1.4cm}@{}}
\toprule
\textbf{Source ID} & \textbf{Source Name} & \textbf{Disease Type} & \textbf{Procedures} & \textbf{Task Type} & \textbf{Pre-train Frames} & \textbf{Evaluation Frames} \\

\midrule

S1 & AVOS \newline (\cite{goodman2024analyzing}) & Multiple & Multiple, Open & Phase/Action/Inst cls.  & 28,473,879 & -- \\
S2 & AIxSuture \newline (\cite{hoffmann2024aixsuture}) & N/A (Skills) & Tissue Suturing & GRS assessment & 2,867,078 & -- \\
S3 & Cholec80 \newline (\cite{yu2018learning}) & Gallbladder  & Cholecystectomy  &  Phase cls., Tool presence & 4,612,530 & -- \\
S4 & CholecT45 \newline (\cite{nwoye2021rendezvous}) & Gallbladder  & Cholecystectomy  & Action cls. (triplets) & 74,855 & -- \\ 

S5 & Colonoscopic & Colorectal lesions & GI diagnosis & Disease cls. & 36,534 & -- \\ %

S6 & Endo-FM \newline (\cite{wang2023foundation}) & Various GI  & Various Endoscopy &  SSL tasks & 3,646,432 & -- \\ 

S7 & SimSurgSkill 2021 &  N/A (Simulation) & Liver/Abdominal (sim) &  Skill metrics, \newline Tool detection & 1,248,156 &  -- \\ \hdashline
S8 & JIGSAWS (2017) \newline (\cite{ahmidi2017dataset}) & N/A (Skills) & Suture/Knot/Needle & Skill assessment, \newline Gesture classification & 569,048 & 537,645 \\ 
S9 & CholecT50 \newline (\cite{nwoye2021rendezvous}) & Gallbladder  & Cholecystectomy  & Action cls., Phase cls., \newline Inst detection & 90,444 & 207,169 \\ 
S10 & AutoLaparo \newline (\cite{wang2022autolaparo}) & Uterine  & Hysterectomy & Phase cls., Motion \newline prediction, Segmentation & 2,155,843 & 160,221 \\
S11 & EndoVis 2019 \newline (\cite{wagner2021comparativevalidationmachinelearning}) & Gallbladder  & Cholecystectomy  &  Phase/Action/Instrument cls., Skill assessment & 4,501,791 & 1,176,000 \\
S12 & Hyper-Kvasir \newline (\cite{Borgli2020}) & GI pathologies & GI procedures & Tissue/Pathology \newline segmentation, Classification & 889,372 & 102,515 \\
S13 & Colonoscopic-web \newline (\cite{mesejo2016computer}) &  Colorectal lesions  & GI diagnosis & Data augmentation & 75,298 & 75,298 \\ 
S14 & Kvasir-Capsule \newline (\cite{Smedsrud2021}) & GI pathologies & GI screening  & Pathology clas.  & 4,765,114 & 61,760 \\
S15 & LDPolypVideo \newline (\cite{ma2021ldpolypvideo}) & Colonic polyps &  Polyp screening &  Polyp detection, \newline Classification & 878,487 & 543,777 \\

S16 & Private Data & Right colon ca.  & Lap. Rt. Hemicolectomy & Phase cls., \newline Skill/Quality assessment & 1,138,833 & 816,952 \\
\midrule
\textbf{Total} & & & && \textbf{56,062,458} & \textbf{3,680,417 } \\
\bottomrule
\end{tabular}}
\caption{\label{tab:surgical_datasets}Summary of the original sources comprising SurgBench Datasets, including publicly available academic datasets, medical competition datasets, demonstration data, and private data. Sources span different disease types, procedures, and task types. Sources above the dashed line do not provide labels and are only used to create SurgBench-P.}
\vspace{0.2cm}
{\footnotesize GI: Gastrointestinal; GRS: Global Rating Score; cls.: classification; sim: simulation; SSL: Self-supervised learning; Obj.: Object; ca.: cancer; Lap.: Laparoscopic; Rt.: Right; Inst.:Instrument}
\end{table*}

We sourced most of the publicly available datasets for surgical video analysis and included 15 public datasets and 1 additional private dataset, expanding 22 surgical procedures, 11 surgical specialties, and relevant tasks, to construct the SurgBench. The full dataset description is shown in Table \ref{tab:surgical_datasets}. Detailed surgical specialties and procedures are provided in Table \ref{tab:surgeries}. In contrast to existing \cite{} work, which usually focuses on specific surgical types or diseases, we unify all of these datasets together, with the goal of building a comprehensive surgical video analysis dataset for more generalizable model performance. However, some datasets in SurgBench are released under licenses that do not allow for secondary distribution. Specifically, AVOS, SimSurgSkill2021, and AIxSuture are under more restrictive licenses (e.g., CC-BY-NC-ND 4.0), which prohibit modification and redistribution. While these datasets were used in pretraining the model, they are not included in the SurgBench-E that we are releasing as the benchmark. Researchers wishing to use these datasets to pretrain the models should request access directly from the dataset providers. Ethical considerations and the original dataset licenses are fully outlined in the appendix \ref{data_license}.

\subsection{Dataset Preprocessing}
Different surgical videos vary in coding format, FPS, video duration, spatial resolution, etc. To standardize data management and accelerate the training process, we perform a series of preprocessing and standardization steps across the 16 video sources. We standardize the videos with H.264 encoding, which offers the best compatibility with various processing libraries, significantly speeding up data loading in the training pipeline. The resolution is compressed to a minimum dimension of 320 (maintaining the aspect ratio, providing space for data augmentation, with the input resolution to the network set at 224x224), and the frame rate is unified to 20-30 FPS.

\subsection{Dataset Formulation}
\label{sec:datasetformulation}
For the format-standardized videos, pretraining and evaluation videos are processed differently. For pretraining data, we employed a fairly straightforward method to process the pre-training data by splitting the unlabelled videos into short clips lasting 10 seconds.  
For evaluation data, we implemented a series of measures to ensure the rationality and usability of benchmark clips. Firstly, we divide all the videos into short clips (lasting 1-10 seconds) with labels. Then, we split them into 5:5 training and testing sets. To prevent the model from focusing on background or scene-specific attributes unrelated to the label \citep{soomro2012ucf101dataset101human}, clips from the same video could only appear in either the training set or the test set. To prevent overly dominant or scarce labels from contaminating the benchmarks, 
\begin{wrapfigure}{r}{0.50\textwidth}
  \centering
  \includegraphics[width=0.50\textwidth]{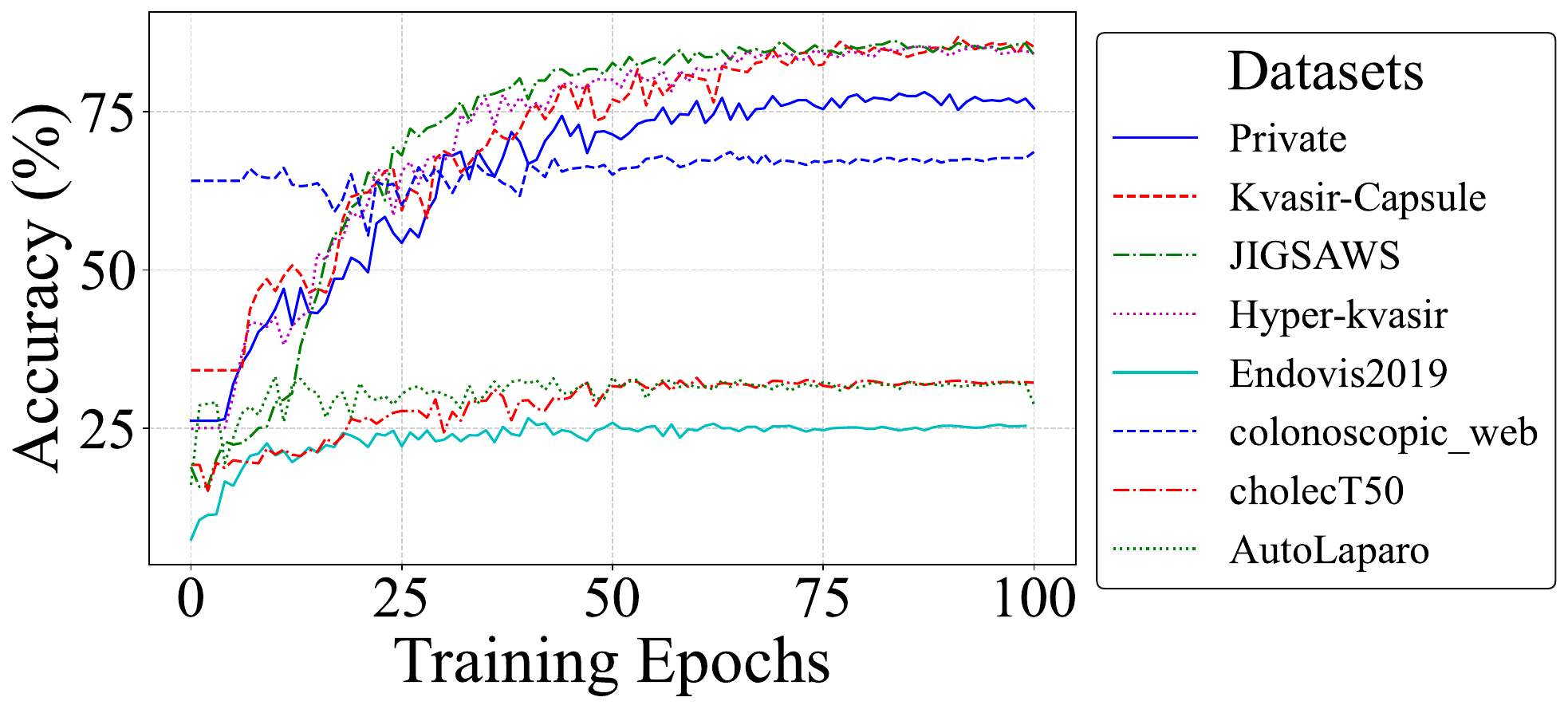}
  \caption{Fine-tuning performance on different source datasets, showing accuracy gains as training progresses. The consistently increasing accuracy curves validate the reliability and data quality of SurgBench-E.}
  \label{fig:dataset valid}
\end{wrapfigure}
the IF (Imbalance Factor) is controlled within 10 through dynamically controlling the clip duration, down-sampling the samples from dominant labels, and removing scarce labels. As for one video corresponding to multiple labels, we split multi-label samples into multiple single-label samples and use the top-k accuracy metric. We have tested the efficacy of each dataset by fine-tuning VideoMAE (standard) on them, and the training dynamics are shown in Figure \ref{fig:dataset valid}. The label distribution of training and testing videos is relatively close, exhibiting a long-tailed distribution, as shown in Figure \ref{fig:train test iid}. The details of processing steps in each dataset are provided in Appendix \ref{details of processing}.

\begin{figure}[htbp]
  \centering
  \includegraphics[width=0.9\textwidth]{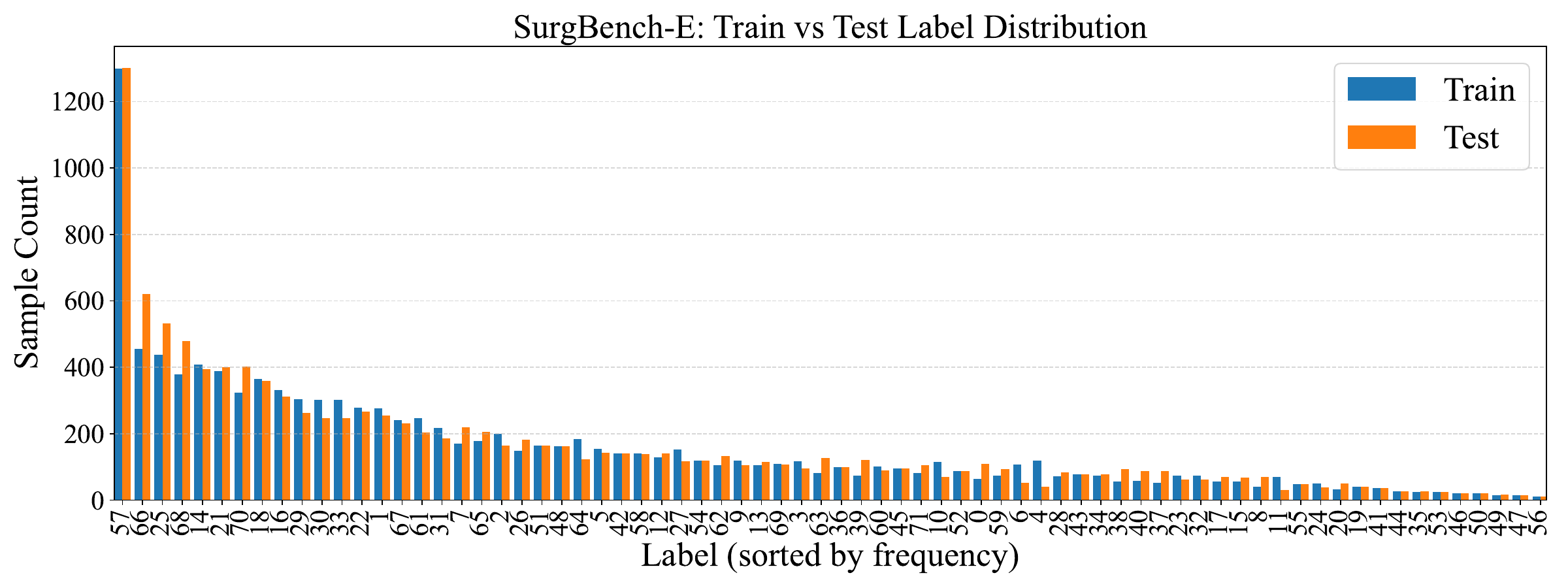}
    \caption{Training and test label distribution of SurgBench-E. The pronounced long-tail pattern aligns with real clinical scenarios, reflecting both authenticity and challenge. Label to task description is illustrated in Table \ref{label index 2 description}.}
  \label{fig:train test iid} 
\end{figure}

\begin{figure}[htbp]
  \centering
  \includegraphics[width=1.0\textwidth]{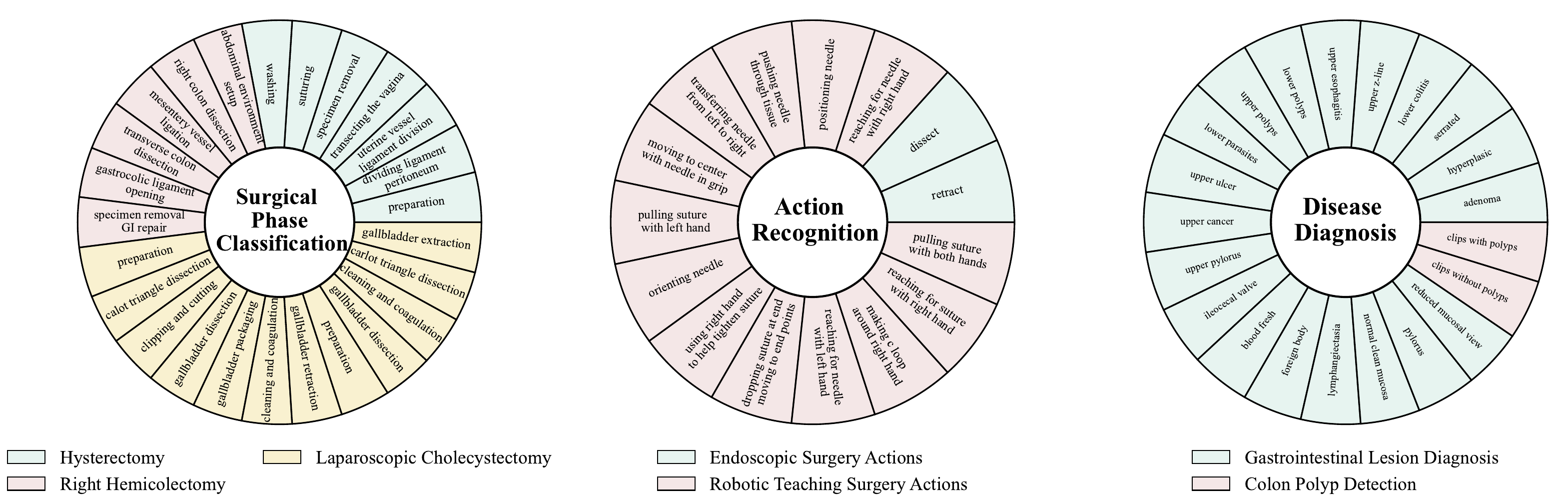}
    \caption{Pie charts of task in SurgBench-E. We have six categories, with three category distributions shown here. Different colors represent different sub-categories. The other three categories are provided in Appendix, Figure \ref{fig:task_labels_others}.}
  \label{fig:task_labels}
\end{figure}

\subsection{Benchmark statistics: }
\begin{wrapfigure}{r}{0.50\textwidth}
  \centering
  \includegraphics[width=0.40\textwidth]{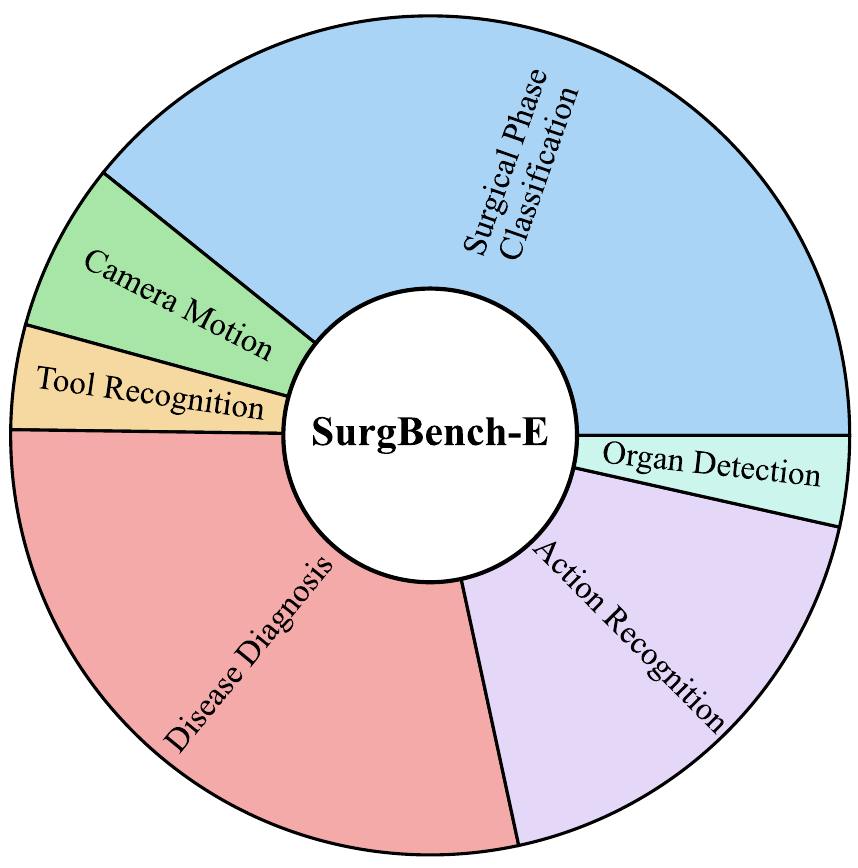}
  \caption{The pie chart of SurgBench-E, containing 6 categories and 72 tasks.}
  \label{fig:6category}
\end{wrapfigure}

The whole SurgBench encompasses 225,250 clips of videos, 59,742,875 frames, lasting 575 hours with 22 surgical procedures and 11 specialties. SurgBench-E contains a total of 23,004 video clips and 72 fine-grained labels for facilitating fine-tuning and testing the model's ability on a wide range of surgical tasks. We define a hierarchical taxonomy called \textbf{6C-10S-72T} on SurgBench-E with \textbf{6} \textbf{C}ategories (three of them are illustrated in Figure \ref{fig:6category}, Figure \ref{fig:task_labels}) and \textbf{10} \textbf{S}ub-Categories with \textbf{72} detailed \textbf{T}asks, as listed in Table \ref{tab:task_codes}. The sample distribution of the 10 sub-categories is naturally long-tailed, as illustrated in Figure \ref{fig:construction pipeline}.The sample size is positively correlated to the number of labels in each category.

\begin{table}[htbp]
\centering
\resizebox{\linewidth}{!}{
\begin{tabular}{llccc}
\toprule
\textbf{Category} & \textbf{Sub-Category (SC)} & \textbf{\# Task} & \textbf{SC Code} & \textbf{Source ID} \\
\midrule
 \multirow{3}{*}{\makecell{ Phase Classification}} & Hysterectomy&7& T1 & S10  \\
  & Right Hemicolectomy &6&T2 & S16\\
  & Laparoscopic Cholecystectomy &12&T3& S9, S11\\\cmidrule(lr){2-2}
 \multirow{1}{*}{\makecell{Camera Motion}} & Camera Motion &7&T4& S10\\ \cmidrule(lr){2-2}
 \multirow{1}{*}{Tool Recognition} & Tool Recognition &2&T5& S9\\\cmidrule(lr){2-2}
 \multirow{2}{*}{Disease Diagnosis} & Gastrointestinal Lesion Diagnosis &19&T6& S12, S13, S14\\
  & Colon Polyp Detection &2&T7& S15\\ \cmidrule(lr){2-2}
 \multirow{2}{*}{Action Classification} & Endoscopic Surgery Actions&2& T8& S9\\
  & Robotic Teaching Surgery Actions &13&T9& S8\\ \cmidrule(lr){2-2}
 \multirow{1}{*}{Organ Detection} & Organ Detection &2&T10& S9\\
\bottomrule
\end{tabular}}
\vspace{5pt}
\caption{The \textbf{6C-10S-72T} taxonomy of our SurgBench-E. Source ID refers to the source of the dataset used in this task. SC codes used in Table~\ref{tab:main table}.}
\label{tab:task_codes}
\end{table}

\section{Methodology}
\label{sec:methodology}
\textbf{Continual pre-training}
We initiated our continual pre-training (CPT) by leveraging VideoMAE models \citep{tong2022videomae} pre-trained on the general-domain Kinetics-400 dataset. The data used for pre-training, SurgBench-P (74.4 million frames total), \textbf{underwent a four-stage refinement} to ensure that it could learn general representations from large-scale data while aligning with downstream tasks. The training steps involved: (1) initial collection of all available videos into 225,250 clips; (2) filtering of less relevant or overly dominant large-scale samples (e.g., from AVOS); (3) applying upsampling for underrepresented data to encourage more IID learning; and (4) a final precise IID-oriented stage with both upsampling and downsampling. This culminated in a refined set of 39,807 video clips (resized to 224$\times$224 resolution) for the final CPT phase. 
Following the VideoMAE methodology, we employed an asymmetric encoder-decoder architecture. The pre-training task involved reconstructing randomly masked spatiotemporal "tubes" of patches using an extremely high masking ratio of 0.9. CPT was performed for two VideoMAE variants: VideoMAE-Standard, which incorporates a ViT-Base encoder and a 4-layer Transformer decoder, and VideoMAE-Large, featuring a ViT-Large encoder paired with a 12-layer Transformer decoder.
CPT parameters for both variants included an input of 16 frames per clip with a temporal sampling rate of 4. We utilized the AdamW optimizer ($\beta_1=0.9, \beta_2=0.95$). The effective batch size was 512 for the Base model (per-GPU batch of 64) and 256 for the Large model (per-GPU batch of 32). The resulting model is named SurgMAE-CPT (Standard version and Large version).

\textbf{Fine-tuning: }
We finetune and test the SurgMAE-CPT on the SurgBench-E, the resulting model is called SurgMAE (standard and large). To validate the effectiveness of continual pre-training on the backbone feature extractor, we opt to train the classifier and analyze the training dynamics. We use AdamW optimizer ($\beta_1=0.9$, $\beta_2=0.99$) with a learning rate of 1e-3, batch size of 64 for fine-tuning standard SurgMAE-CPT, whereas the large SurgMAE-CPT is trained with a learning rate of 2e-3, batch size of 16.

\textbf{Training cost: } For continual pre-training, the experiments were conducted on a cluster of 8 NVIDIA A100 GPUs. The VideoMAE-Base model was continually pre-trained for a total of 54 epochs on the surgical data, while the VideoMAE-Large model was trained for 38 epochs. The CPT process for both versions collectively took one week. For fine-tuning, we use one host with 8 RTX3090 (24 GB) GPUs. Both standard and large models are trained for 100 epochs, lasting 14 hours to converge.

\section{Experiments and Results}
\label{sec:experiments}

In this section, we present the experimental setup and results to validate the utility of SurgBench for surgical video analysis. We evaluate the performance of models trained using self-supervised pretraining with VideoMAE and fine-tuned on SurgBench-E. Our experiments focus on understanding training dynamics, model scalability, the impact of hyperparameters, generalization capabilities, and the benefits of continued pretraining.

\subsection{Main Experiment Results}

\begin{table}[t]
\centering
\begin{adjustbox}{width=0.85\textwidth}
\begin{tabular}{c|cc|cc|cc|cc}
\toprule
\multirow{2}{*}{Task} & \multicolumn{2}{c|}{Random Init.} & \multicolumn{2}{c|}{VideoMAE} & \multicolumn{2}{c|}{SurgMAE} & \multicolumn{2}{c}{SurgMAE(L)} \\
\cmidrule(lr){2-9}
 & Top-1 & Top-3 & Top-1 & Top-3 & Top-1 & Top-3 & Top-1 & Top-3 \\
\midrule
T1  & 0.000 & 0.162 & 0.125 & 0.356 & 0.185 & 0.484 & \textbf{0.182} & \textbf{0.501} \\
T2  & 0.364 & 0.667 & 0.423 & 0.777 & 0.440 & 0.818 & \textbf{0.493} & \textbf{0.856} \\
T3  & 0.126 & 0.523 & 0.379 & 0.653 & 0.399 & 0.671 & \textbf{0.422} & \textbf{0.742} \\
T4  & 0.000 & 0.000 & 0.298 & 0.629 & 0.293 & 0.628 & \textbf{0.306} & \textbf{0.645} \\
T5  & 0.395 & \textbf{0.799} & 0.287 & 0.766 & 0.319 & 0.749 & \textbf{0.324} & 0.760 \\
T6  & 0.413 & 0.605 & 0.569 & 0.744 & 0.645 & 0.838 & \textbf{0.666} & \textbf{0.905} \\
T7  & 0.003 & 0.045 & 0.374 & 0.534 & 0.561 & 0.791 & \textbf{0.739} & \textbf{0.930} \\
T8  & 0.000 & 0.044 & 0.005 & 0.254 & 0.015 & 0.383 & \textbf{0.020} & \textbf{0.411} \\
T9  & 0.003 & 0.408 & 0.102 & 0.446 & 0.475 & 0.743 & \textbf{0.627} & \textbf{0.825} \\
T10 & 0.000 & 0.290 & 0.122 & 0.632 & 0.200 & 0.575 & \textbf{0.279} & \textbf{0.695} \\
\midrule
Avg. & 0.229 & 0.478 & 0.378 & 0.652 & 0.448 & 0.731 & \textbf{0.487} & \textbf{0.788} \\
\bottomrule
\end{tabular}
\end{adjustbox}
\vspace{5pt}
\caption{Top-1 and Top-3 accuracy of models across tasks (T1–T10). Bold indicates best performance per task. The model is standard-sized except (L), which stands for large.}
\label{tab:main table}
\end{table}

We conducted pre-training using the VideoMAE architecture on the SurgBench-P dataset (with all parameters fine-tuned) and further fine-tuned on SurgBench-E. The performance across 10 sub-categories is summarized in Table \ref{tab:main table}. To mitigate the interference caused by multiple categories present in a single sample, which can lead to unstable results, we considered both top-1 accuracy and top-3 accuracy. To compare the representational capacity of different backbones, we froze the feature extractor during fine-tuning and only trained the classification head.

The results show that continual pre-training on surgical videos has significantly improved accuracy on SurgBench-E, with consistent growth observed across metrics. Specifically, top-1 accuracy increased by 7\% and top-3 accuracy by 7.9\%. Most of the other sub-categories also showed noticeable improvement, except for T4 and T8, where the performance remained nearly unchanged. This suggests that the pre-training data has reliable quality and that self-supervised pre-training helps learn robust representations.

\begin{figure}[h]
  \centering
  \begin{subfigure}[b]{0.44\textwidth}
    \centering
    \includegraphics[width=\textwidth]{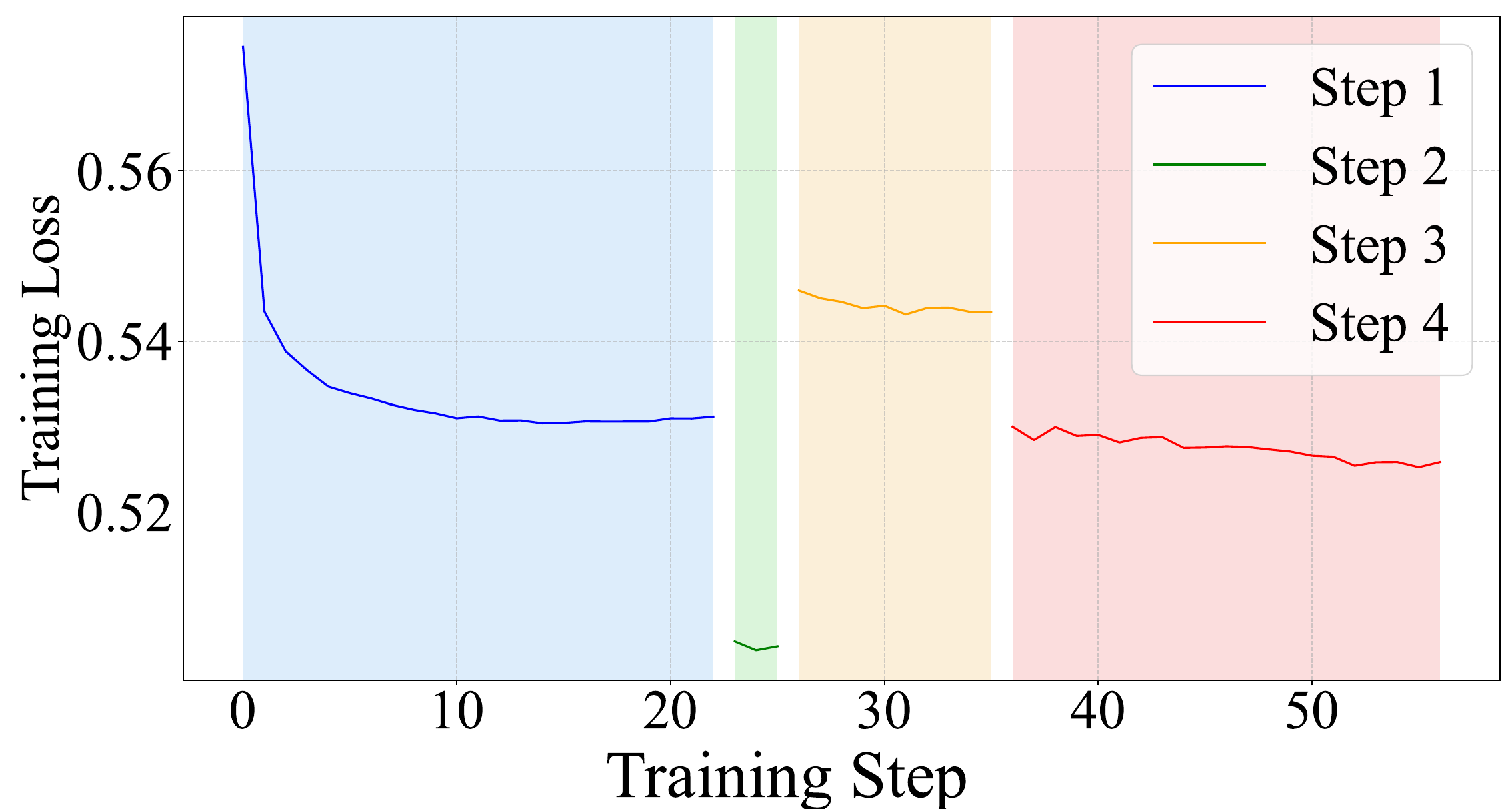}
    \caption{Continuing pretraining dynamics of standard SurgMAE on SurgBench-P across 4 steps.}
  \end{subfigure}
  \hfill
  \begin{subfigure}[b]{0.44\textwidth}
    \centering
    \includegraphics[width=\textwidth]{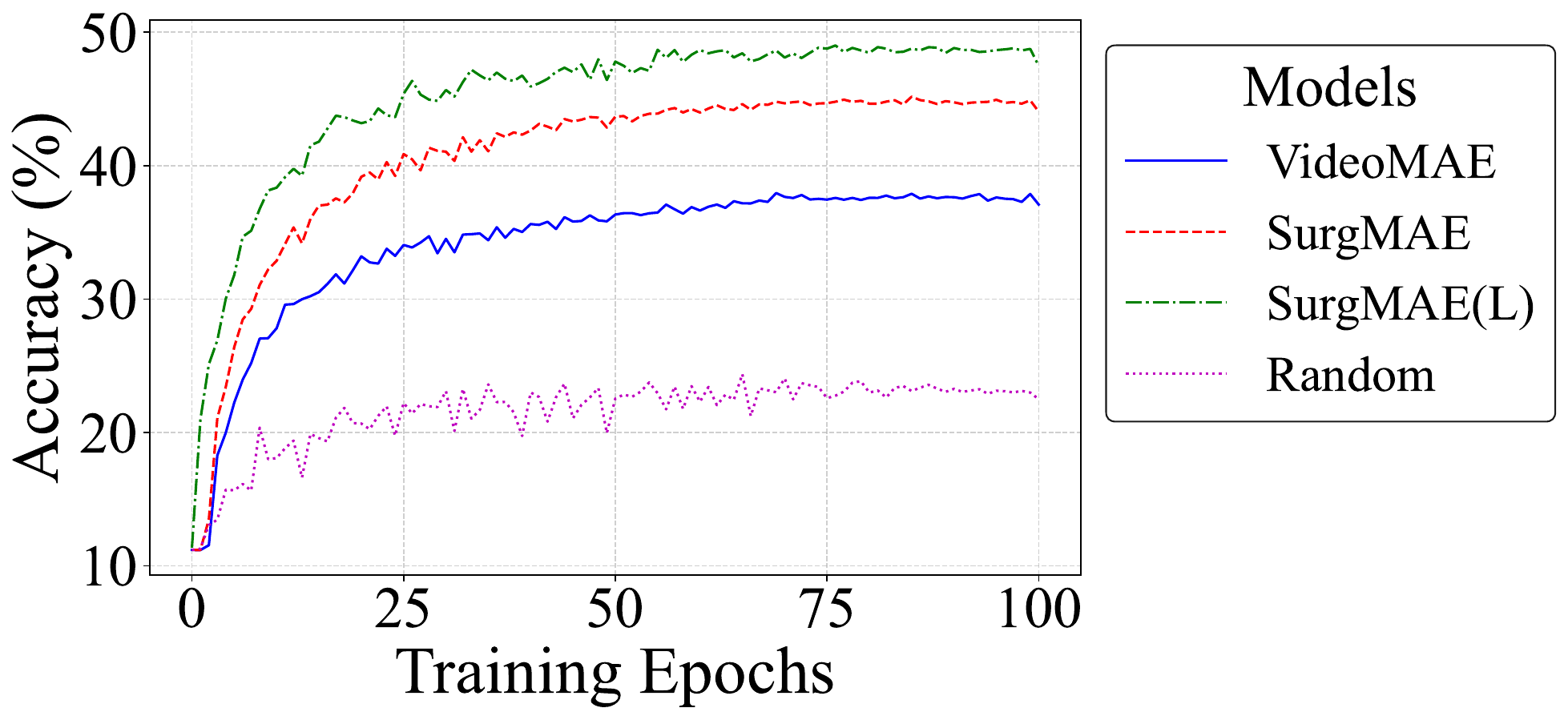}
    \caption{Fine-tuning dynamics of different models on SurgBench-E. The model is standard-sized except (L), which stands for large version.}
  \end{subfigure}
    \caption{Dynamics of pretraining and fine-tuning on SurgBench. For pretraining, we track 4 steps progressing from large-scale data to small-scale distribution alignment with SurgBench-E. For fine-tuning, we compare randomly initialized weights, SurMAE pretrained on Kinetics, and large/standard versions pretrained on SurgBench-P. The stable decrease in loss curves and increase in accuracy demonstrate the rationality and quality of our dataset.}
  \label{fig:training_dynamics}
\end{figure}

To gain a detailed understanding of which specific categories exhibit performance fluctuations on our pre-training and small task sets, we plotted the performance metrics for each category, as shown in Table~\ref{tab:per_class_performance}. We identified two key patterns. First, the performance for a subset of categories is notably low. This is attributed to the scarcity of certain rare types in a few video domains, resulting in extremely limited data volume. This presents a challenge for future model optimization: how to achieve high performance for categories that are narrowly defined or underrepresented. Second, the pre-training process consistently improves the performance across nearly all individual categories, demonstrating the model's robust generalization capability.

\subsection{Training Dynamics}

A stable training loss curve can strongly imply the quality of the data and the effectiveness of the training scheme. During pre-training, to evaluate the effectiveness of the data and training strategy, we analyzed the loss dynamics across four stages of progressive data distribution in the VideoMAE standard version. We find that within each stage, the loss decreases very smoothly, as seen in Figure~\ref{fig:training_dynamics}. Moreover, due to differences in the difficulty and distribution of the training data, the transition of the loss values between different stages is not smooth. However, in actual task testing, this lack of smoothness does not negatively affect performance. 

Furthermore, we observe that even though the loss curve in the final stage does not show a significant reduction, it still contributes to gradually stabilizing and improving the performance of task fine-tuning. This indicates that the model can continue representation learning even when the loss decreases more slowly.

\begin{wrapfigure}[18]{r}{0.45\textwidth}
  \centering
  \includegraphics[width=0.45\textwidth]{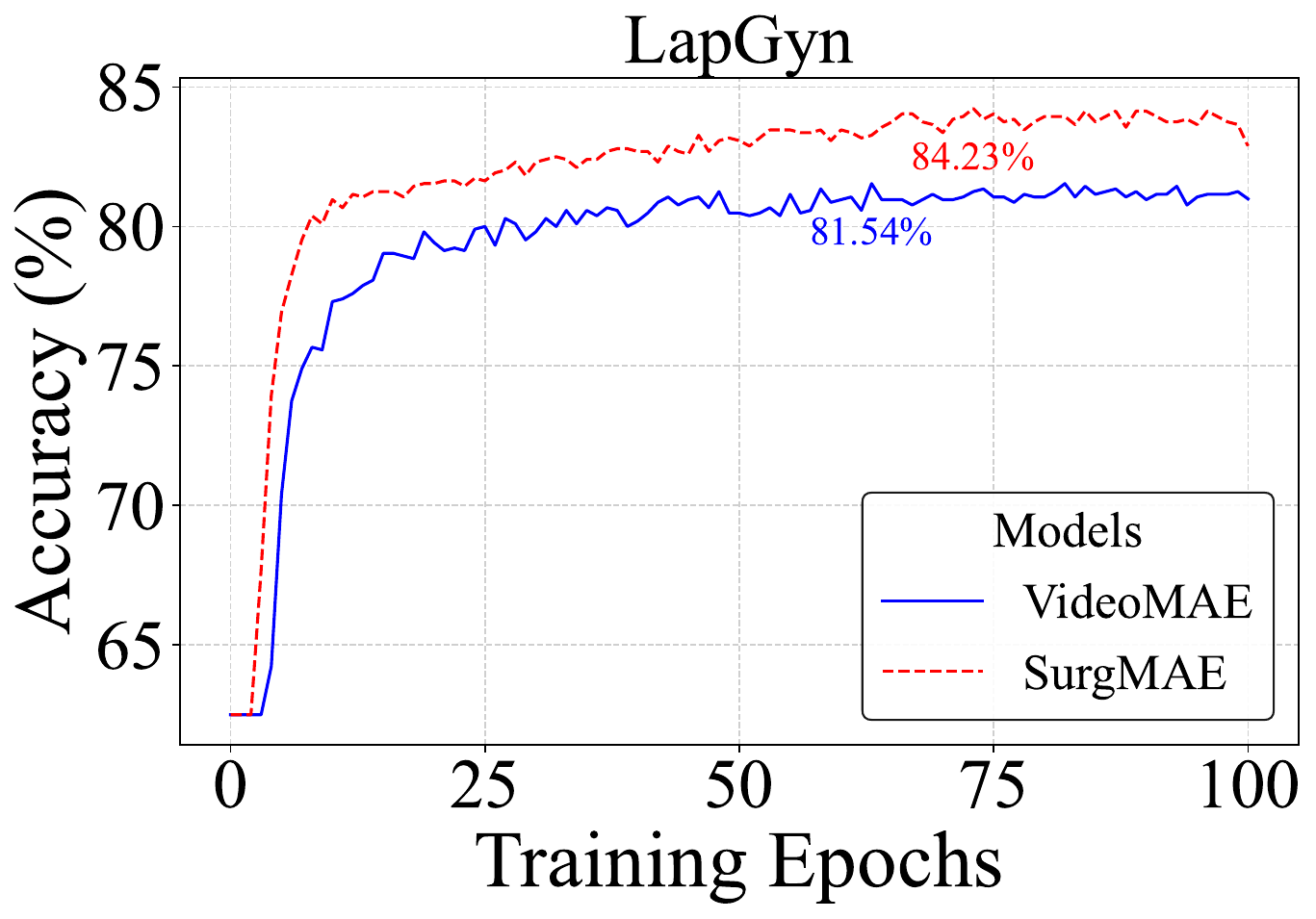}
  \caption{Performance on OOD dataset. We tested on the LapGyn task, which contains out-of-distribution data not seen during SurgBench-P training. The pre-trained model demonstrates superior convergence speed and final accuracy when fine-tuned on this out-of-domain downstream task.}
  \label{fig:ood}
\end{wrapfigure}
Similarly, we also examined the quality of downstream task labels through training dynamics. We found that as the number of training epochs increases, the accuracy steadily rises. This phenomenon is consistent across the standard model, the large parameter model, the Kinetics pre-trained model, and the randomly initialized model.

\subsection{Impact of Model Size on Performance}
In language pre-training models, larger models with more parameters and scale typically exhibit better performance and stronger generalization capabilities. In the context of surgical foundation models, we also tested the performance of standard-sized and larger models on downstream tasks. As shown in Table 4, SurgMAE and SurgMAE(L) achieved a 3.9\% increase in top-1 accuracy and a 5.7\% increase in top-3 accuracy. 

Furthermore, consistent growth was observed across task dimensions. This phenomenon is in line with the behavior of large-scale language models. Additionally, subsequent work on surgical procedure understanding suggests that further increasing the model scale can lead to even higher performance.

\subsection{Can Surg-FM generalize to unseen case?}
Models trained on SurgBench can improve generalization across diverse distributions. We evaluated performance on datasets with varying degrees of similarity to SurgBench. Specifically, we tested on LapGyn4, a comprehensive dataset from gynecologic laparoscopic surgeries categorized into four distinct tasks (surgical actions, anatomical structures, actions on anatomy, and instrument count) \citep{DBLP:conf/mmsys/LeibetsederPPKM18}. We focused on an event recognition task in Laparoscopic Gynecology as described in \citep{nasirihaghighi2024event}. The fine-tuning dynamics are shown in Figure \ref{fig:ood}. Remarkably, models pretrained on SurgBench demonstrated significant performance gains of 2.69\% even on this dissimilar distribution.


\subsection{Data mix strategy between Pre-training and  Fine-tuning  }
\begin{figure}[h]
  \centering
  \begin{subfigure}[b]{0.45\textwidth}
    \centering
    \includegraphics[width=\textwidth]{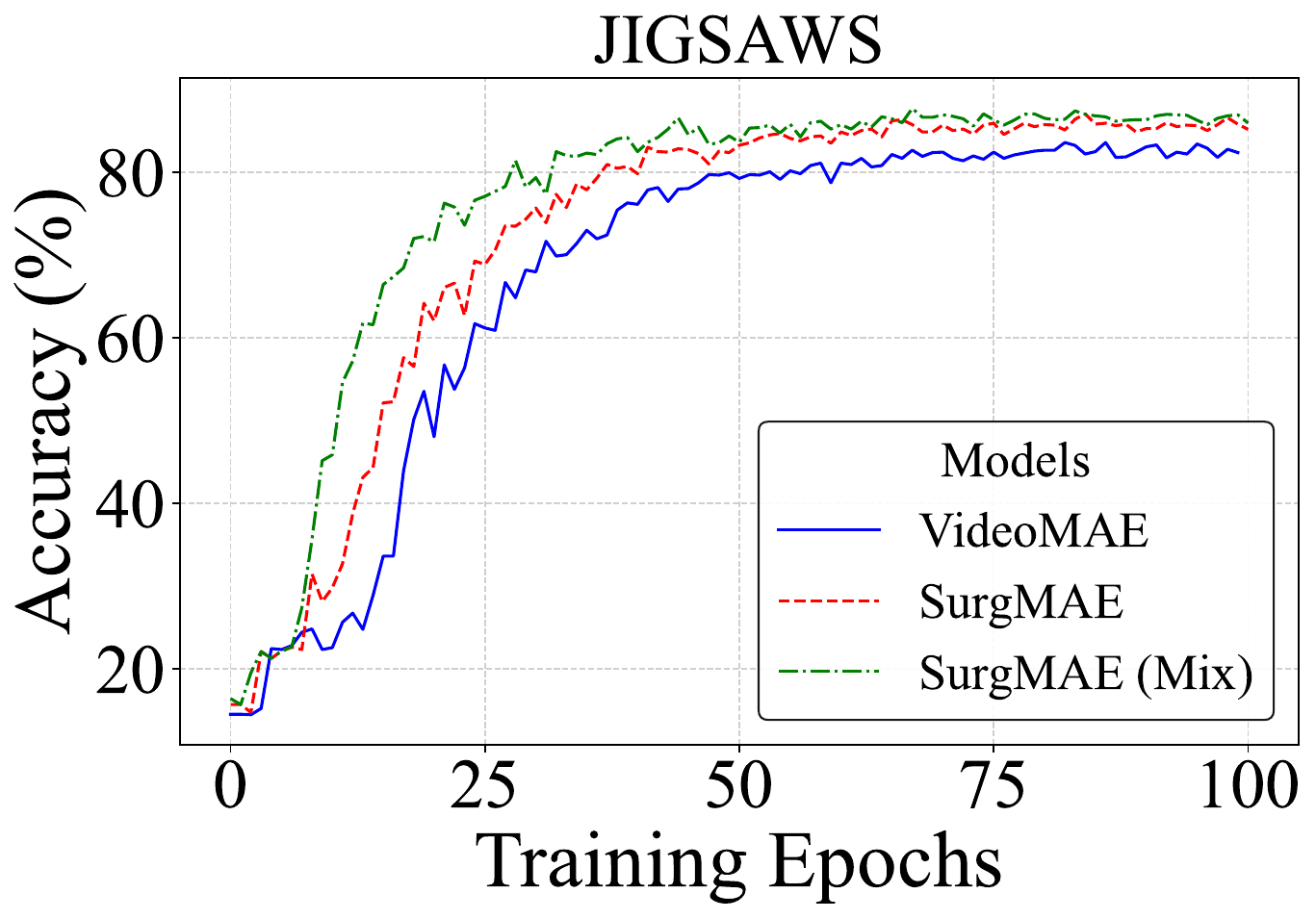}
    \caption{Fine-tuning dynamics of SurgMAE pretrained on mixed data. We first pretrained the model using a combined dataset of JIGSAWS and CholecT50, then fine-tuned it on JIGSAWS alone.}
  \end{subfigure}
  \hfill
  \begin{subfigure}[b]{0.45\textwidth}
    \centering
    \includegraphics[width=\textwidth]{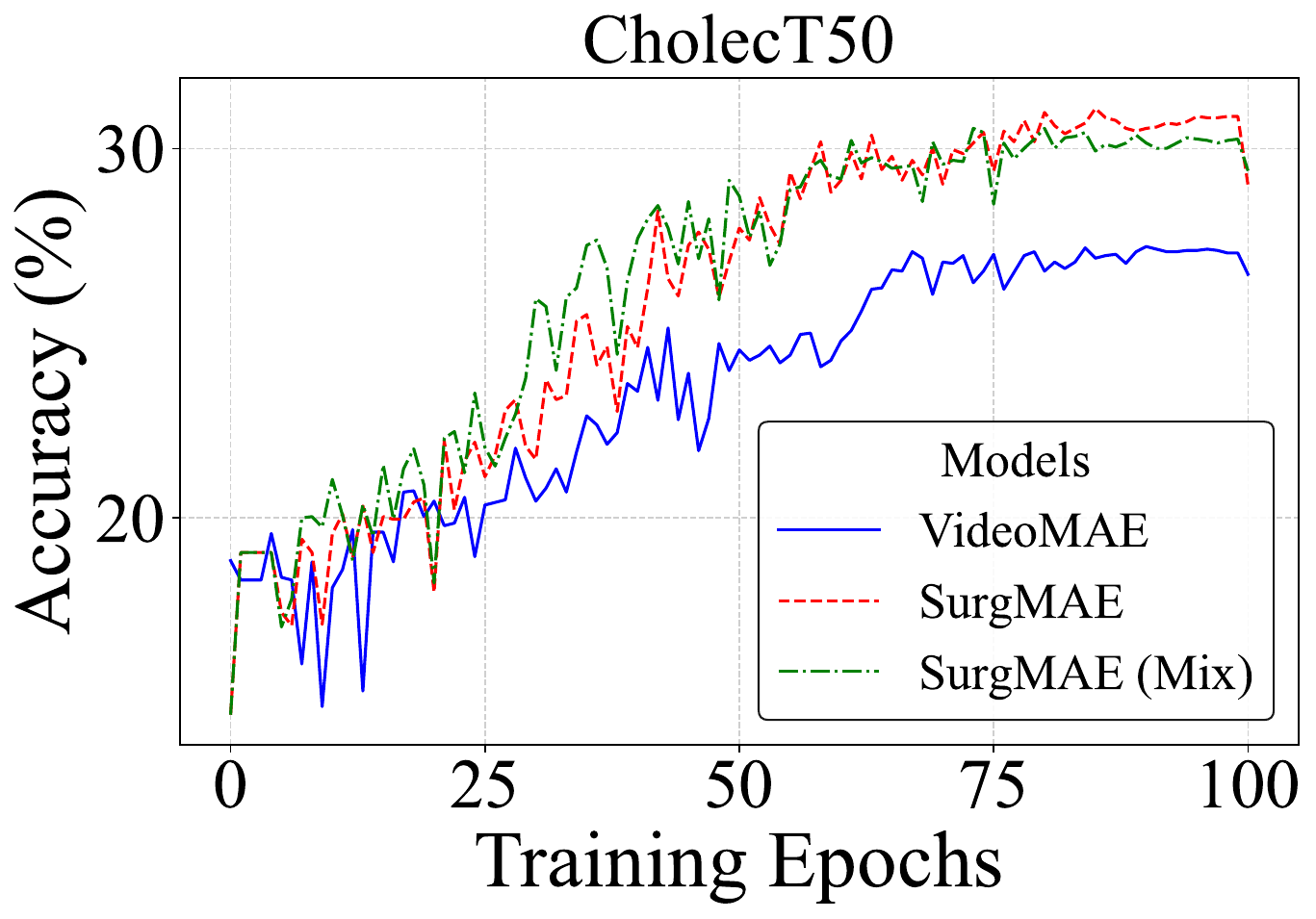}
    \caption{Fine-tuning dynamics of SurgMAE on CholecT50. After pretraining on combined JIGSAWS and CholecT50 data, the model was fine-tuned specifically on the CholecT50 dataset.}
  \end{subfigure}
  \caption{Comparison of fine-tuning performance between models pretrained on mixed surgical data versus models trained from scratch, demonstrating the effectiveness of our data mixing strategy.}
  \label{fig:combined_dynamics}
\end{figure}

Data mixing during training significantly impacts downstream task performance. In SurgBench-P, we combined all source data for pre-training and used a strategy transitioning from "large-scale + unrestricted distribution" to "small-scale + aligned distribution." We hypothesize that optimal performance will follow as long as the training distribution aligns with the downstream task distribution in the final stage. Using the dataset (JIGSAWS and CholecT50) for both continuing pre-training on Kinetics-400 and fine-tuning, we observed that our model achieved superior convergence speed and final performance, seen in Figure~\ref{fig:combined_dynamics}. 

\vspace{-5pt}
\section{Conclusions}
\label{sec:conclusion}
We present SurgBench, a comprehensive benchmark for surgical video understanding. Our contributions include: (1) SurgBench-P, a diverse pretraining dataset comprising 53 million frames across 22 surgical procedures and 11 surgical specialties; (2) SurgBench-E, a structured evaluation benchmark with 6 category, 10 sub-categories, and 72 fine-grained tasks, facilitating comprehensive surgical video benchmarking; and (3) empirical validation demonstrating significant performance gains through self-supervised learning on surgical videos. Experimental results show that models pretrained on SurgBench outperform those pretrained on natural video datasets. SurgBench provides a unified platform for surgical video analysis that will accelerate research progress in this domain.

\section{Limitations}
Despite its contributions, SurgBench has several limitations: (1) the long-tail class distribution challenges model training, requiring further research to improve minority class performance; (2) language supervision integration remains unexplored but potentially beneficial; and (3) optimal architectural designs for surgical video foundation models require additional investigation to maximize cross-task performance.



{\small
\bibliographystyle{apalike}  
\bibliography{egbib}    
}


\appendix

\section{Dataset License}
\label{data_license}
SurgBench aggregates several publicly available surgical video datasets, with their respective licenses meticulously reviewed to ensure compliant use within the research community. The core components of SurgBench, intended for direct benchmarking and fine-tuning, primarily consist of datasets released under licenses permitting academic and non-commercial research, such as Cholec80 (CC-BY-NC-SA 4.0), Kvasir-Capsule (custom academic/educational use), and SUN-SEG (custom non-commercial research use over Apache License). Notably, while foundational models developed alongside SurgBench were pre-trained using a broader corpus that included datasets with more restrictive terms like AVOS, SimSurgSkill2021, and AIxSuture (CC-BY-NC-ND 4.0), these specific datasets are not redistributed as part of SurgBench's fine-tuning suite due to their licensing limitations (e.g., challenge-specific, no derivatives). Users are strongly encouraged to consult the original sources for comprehensive licensing details of each constituent dataset.

\begin{table}[htbp]
\centering
\footnotesize
\begin{tabular}{|l|l|l|}
\hline
\textbf{Dataset Name} & \textbf{License Type} & \textbf{Usage Conditions} \\
\hline
\multicolumn{3}{|c|}{\textbf{Datasets for SurgBench-P and SurgBench-E}} \\
\hline
Cholec80 & CC-BY-NC-SA 4.0 & Attribution required, non-commercial use, share-alike \\
\hline
Hyper-Kvasir & Custom & Citation required, academic use \\
\hline
Kvasir-Capsule & Custom & Attribution required, research and educational purposes \\
\hline
Colonoscopic Dataset & Custom & Publicly available, registration required \\
\hline
AutoLaparo & CC-BY-NC-SA 4.0 & Attribution required, non-commercial use, share-alike \\
\hline
SUN-SEG & Custom & Application required, research and educational purposes \\
\hline
CholecT50 & CC-BY-NC-SA 4.0 & Attribution required, non-commercial use, share-alike \\
\hline
JIGSAWS & Custom & Application required, academic research, specific citations \\
\hline
Endoscopic Vision 2019 & Custom & Form submission required, academic use \\
\hline
\multicolumn{3}{|c|}{\textbf{Datasets For SurgBench-P Only}} \\
\hline
SimSurgSkill2021 & Challenge-specific & Limited to challenge scope only \\
\hline
AVOS & Custom protocol & Redistribution prohibited, academic research only \\
\hline
AIxSuture & CC-BY-NC-ND 4.0 & Modifications and commercial use prohibited \\
\hline
\end{tabular}
\caption{Classification of Surgical Datasets by License}
\label{tab:dataset_licenses}
\end{table}

\section{Pie chart of other three categories}
\begin{figure}[htbp]
  \centering
  \includegraphics[width=1.0\textwidth]{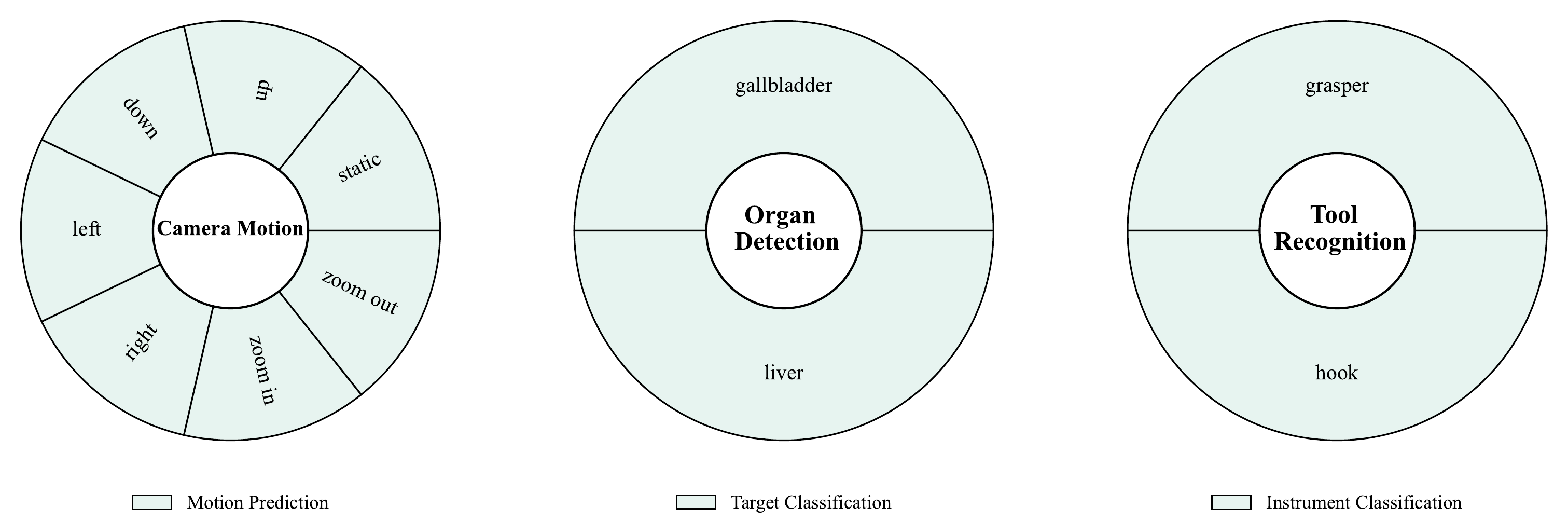}
    \caption{Pie charts of task in SurgBench-E. We have six categories, with three category distributions shown here. Different colors represent different sub-categories.}
  \label{fig:task_labels_others}
\end{figure}

The pire chart of the camera motion, organ detection, tool recognition is shown in Figure \ref{fig:task_labels_others}.

\section{Pretraining Data Composition}
The composition of pretraining data is shown in Table \ref{tab:surgeries},
\begin{table}[htbp]
\centering
\begin{tabular}{|l|l|l|}
\hline
\textbf{Clinical Specialty} & \textbf{Surgical Procedure} & \textbf{Source ID} \\ \hline
General Surgery & Appendectomy & S1, S6 \\ \hline
General Surgery & Cholecystectomy & S1, S3, S4, S9, S11 \\ \hline
General Surgery & Gastrectomy & S1, S6 \\ \hline
General Surgery & Hernia Repair & S1, S6 \\ \hline
General Surgery & Splenectomy & S1, S7 \\ \hline
General Surgery & Pilonidal cystectomy & S1 \\ \hline
General Surgery & Bariatric surgery & S1, S6 \\ \hline
General Surgery & Anti-reflux surgery & S1, S12 \\ \hline
General Surgery & Hepatectomy & S1, S6, S7 \\ \hline
Colon and Rectal Surgery & Colectomy & S1, S6, S12, S16 \\ \hline
Colon and Rectal Surgery & Colostomy & S1, S6, S12 \\ \hline
Thoracic Surgery & Esophagectomy & S1, S6, S12 \\ \hline
Urological Surgery & Nephrectomy & S1 \\ \hline
Urological Surgery & Adrenalectomy & S1 \\ \hline
Obstetrics and Gynecology & Hysterectomy & S1, S10 \\ \hline
Obstetrics and Gynecology & Gynecologic Oncology & S1, S10 \\ \hline
Neurological Surgery & Neurological Surgery & S1 \\ \hline
Ophthalmic Surgery & Ophthalmic Surgery & S1 \\ \hline
Oral and Maxillofacial Surgery & Oral and Maxillofacial Surgery & S1 \\ \hline
Orthopaedic Surgery & Orthopaedic Surgery & S1 \\ \hline
Otolaryngology & Otolaryngology & S1 \\ \hline
Pediatric Surgery & Pediatric Surgery & S1 \\ \hline
\end{tabular}
\caption{Clinical specialties and surgical procedures and their sources (ID) of SurBennch-P}
\label{tab:surgeries}
\end{table}

\section{Label to task mapping}
The 72 label to task mapping is shown in Table \ref{label index 2 description}, \ref{label2description2}.
\begin{table}[ht]
\centering
\begin{tabularx}{\textwidth}{|c|X|}
\hline
Label & Description \\
\hline
0 & surgical\_phase-preparation \\
1 & surgical\_phase-dividing\_ligament\_and\_peritoneum \\
2 & surgical\_phase-dividing\_uterine\_vessels\_and\_ligament \\
3 & surgical\_phase-transecting\_the\_vagina \\
4 & surgical\_phase-specimen\_removal \\
5 & surgical\_phase-suturing \\
6 & surgical\_phase-washing \\
7 & motion\_prediction-static \\
8 & motion\_prediction-up \\
9 & motion\_prediction-down \\
10 & motion\_prediction-left \\
11 & motion\_prediction-right \\
12 & motion\_prediction-zoom-in \\
13 & motion\_prediction-zoom-out \\
14 & instrument\_classification-grasper \\
15 & instrument\_classification-hook \\
16 & verb\_classification-retract \\
17 & verb\_classification-dissect \\
18 & target\_classification-gallbladder \\
19 & target\_classification-liver \\
20 & phase\_classification-preparation \\
21 & phase\_classification-carlot-triangle-dissection \\
22 & phase\_classification-gallbladder-dissection \\
23 & phase\_classification-cleaning-and-coagulation \\
24 & phase\_classification-gallbladder-extraction \\
25 & disease\_classification-adenoma \\
26 & disease\_classification-hyperplasic \\
27 & disease\_classification-serrated \\
28 & phase\_classification-preparation \\
29 & phase\_classification-calot\_triangle\_dissection \\
30 & phase\_classification-clipping\_and\_cutting \\
31 & phase\_classification-galbladder\_dissection \\
32 & phase\_classification-galbladder\_packaging \\
33 & phase\_classification-cleaning\_and\_coagulation \\
34 & phase\_classification-galbladder\_retraction \\
35 & disease\_classification-home \\
36 & disease\_classification-home \\
37 & disease\_classification-home \\

\hline
\end{tabularx}
\caption{Label Descriptions (Part 1) \label{label index 2 description}}
\end{table}

\vspace{0.5cm}

\begin{table}[ht]
\centering
\begin{tabularx}{\textwidth}{|c|X|}
\hline
Label & Description \\
\hline
38 & disease\_classification-home \\
39 & disease\_classification-home \\
40 & disease\_classification-home \\
41 & disease\_classification-home \\
42 & disease\_classification-home \\
43 & disease\_classification-home \\
44 & gesture\_classification-reaching\_for\_needle\_with\_right\_hand \\
45 & gesture\_classification-positioning\_needle \\
46 & gesture\_classification-pushing\_needle\_through\_tissue \\
47 & gesture\_classification-transferring\_needle\_from\_left\_to\_right \\
48 & gesture\_classification-moving\_to\_center\_with\_needle\_in\_grip \\
49 & gesture\_classification-pulling\_suture\_with\_left\_hand \\
50 & gesture\_classification-orienting\_needle \\
51 & gesture\_classification-using\_right\_hand\_to\_help\_tighten\_suture \\
52 & gesture\_classification-dropping\_suture\_at\_end\_and\_moving\_to\_end\_points \\
53 & gesture\_classification-reaching\_for\_needle\_with\_left\_hand \\
54 & gesture\_classification-making\_c\_loop\_around\_right\_hand \\
55 & gesture\_classification-reaching\_for\_suture\_with\_right\_hand \\
56 & gesture\_classification-pulling\_suture\_with\_both\_hands \\
57 & disease\_classification-ileocecal\_valve \\
58 & disease\_classification-blood-fresh \\
59 & disease\_classification-foreign-body \\
60 & disease\_classification-lymphangiectasia \\
61 & disease\_classification-normal-clean-mucosa \\
62 & disease\_classification-pylorus \\
63 & disease\_classification-reduced-mucosal-view \\
64 & polyp\_detection-clips\_without\_polyps \\
65 & polyp\_detection-clips\_with\_polyps \\
66 & phase\_classification-1\_establishment\_of\_abdominal\_surgical\_environment \\
67 & phase\_classification-2\_dissection\_of\_the\_posterior\_peritoneal\_space \\
68 & phase\_classification-3\_\_identification\_and\_ligation\_of\_the\_vessels\_on\_the\_mesentery \\
69 & phase\_classification-4\_dissection\_of\_the\_posterior\_peritoneal\_space \\
70 & phase\_classification-5\_opening\_of\_the\_gastrocolic\_ligament\_and\_identification \\
71 & phase\_classification-6\_specimen\_removal\_and\_gastrointestinal\_reconstruction\_(intra-abdominal\_and\_extra-abdominal\_anastomosis) \\
\hline
\end{tabularx}
\caption{Label Descriptions (Part 2)}\label{label2description2}
\end{table}
\section{Details of processing steps in each dataset}\label{details of processing}

\noindent\textbf{Private Data}
The dataset used in this study is derived from laparoscopic right hemicolectomy procedures and focuses on surgical phase classification. The original dataset consists of surgical videos collected from \textbf{three tertiary grade A hospitals} in China. These videos were initially stored in their raw format, organized under a hierarchical directory structure corresponding to different surgical phases as outlined in the Competency Assessment Tool (CAT) for laparoscopic right hemicolectomy. The primary data components include six distinct phases: 1. Establishment of abdominal surgical environment, 2. Dissection of the posterior peritoneal space of the right colon and the right hemi-colon, 3. Identification and ligation of the vessels on the mesentery, 4. Dissection of the posterior peritoneal space of the transverse colon and the Henle's trunk, 5. Opening of the gastrocolic ligament and identification of the mesenteric interspace (IMS), and 6. Specimen removal and gastrointestinal reconstruction (intra-abdominal and extra-abdominal anastomosis). Each video was annotated with detailed metadata regarding exposure, adverse events, technical operations, and quality evaluations, enabling a comprehensive assessment of surgical proficiency.

To prepare the dataset for machine learning tasks, we implemented a systematic preprocessing pipeline using Python scripts. The raw videos were segmented into 30-second clips, which were then accelerated by a factor of 3 to compress each clip into 10 seconds while preserving the essential visual information. This process utilized the \texttt{ffmpeg} library, ensuring precise frame-level extraction and efficient compression. The segmentation logic involved calculating the total video duration using \texttt{ffprobe}, dividing it into non-overlapping intervals, and applying a single filter chain for both video and audio streams to maintain synchronization. Clips were saved in the H.264 codec with AAC audio encoding, optimized for fast network streaming using the \texttt{+faststart} flag. Metadata for each clip, including its label, relative path, compressed duration, and associated task type, was recorded in a JSON file for subsequent model training. 

After processing, the dataset was divided into $N$ clips across the six categories, with each clip having a standardized duration of approximately 10 seconds. The resolution of the videos was preserved at $1920 \times 1080$ pixels, with a frame rate of 30 frames per second (fps). The total number of frames across all clips amounted to approximately $30 \times N$, providing a rich resource for temporal modeling. The private data, is structured hierarchically, with each clip labeled according to its surgical phase, enabling supervised learning approaches for phase recognition. The final metadata file, containing detailed annotations for each clip, was saved in JSON format.

\noindent\textbf{AutoLaparo}
The AutoLaparo is a multi-task dataset specifically designed for advancing image-guided surgical automation in laparoscopic hysterectomy. The original dataset comprises three main tasks: surgical workflow recognition (Task 1), laparoscope motion prediction (Task 2), and instrument and key anatomy segmentation (Task 3). Task 1 includes 21 laparoscopic hysterectomy videos, each annotated at 1 fps with seven surgical phases, while Task 2 consists of 300 video clips, each lasting 10 seconds, annotated with seven laparoscopic motion labels. Task 3 provides 1800 images with corresponding masks, annotated for four types of surgical instruments and one key anatomy. The dataset is organized into specific directories: Task 1 videos and labels are stored in \texttt{task1/videos} and \texttt{task1/labels}, respectively; Task 2 clips and their labels are located in \texttt{task2/clips} and \texttt{task2/laparoscope\_motion\_label.txt}; and Task 3 images and masks are stored in \texttt{task3/images} and \texttt{task3/masks}. For preprocessing, Task 1 videos are segmented into clips of 100 frames each, ensuring a consistent clip length, while Task 2 clips are directly used as provided. The dataset is split into training, validation, and testing sets: for Task 1, videos 01-10 are used for training, 11-14 for validation, and 15-21 for testing; for Task 2 and Task 3, clips 001-170 are designated for training, 171-227 for validation, and 228-300 for testing. Post-processing, the dataset contains 1,388 minutes of video, with a frame rate of 25 fps and a resolution of 1920$\times$1080 pixels. This meticulous division ensures a balanced and representative dataset for robust model training and evaluation, facilitating comprehensive benchmarking in surgical automation tasks.

\noindent\textbf{Cholec80}
The Cholec80 dataset, originating from the University Hospital of Strasbourg/IRCAD, is a comprehensive collection of laparoscopic cholecystectomy videos designed for surgical workflow analysis. This dataset includes 80 high-resolution videos, each annotated with detailed phase labels and tool presence information, making it a valuable resource for research in surgical video understanding. The dataset's homepage (\url{http://camma.u-strasbg.fr/datasets}) provides an overview of its applications, including phase recognition and tool detection tasks. Additionally, the accompanying README file and annotations offer insights into the data structure, licensing, and citation requirements.

Initially, the dataset comprises 80 videos recorded at 25 frames per second (fps), with resolutions varying across videos. Each video is accompanied by two types of annotations: phase annotations and tool annotations. The phase annotations provide frame-level labels for seven surgical phases, including preparation, Calot triangle dissection, clipping and cutting, gallbladder dissection, gallbladder packaging, cleaning and coagulation, and gallbladder retraction. Tool annotations, sampled at 1 fps, indicate the presence or absence of seven surgical tools, such as graspers, bipolar devices, hooks, scissors, clippers, irrigators, and specimen bags. These annotations are stored in tab-separated text files, with frame indices and corresponding labels. Furthermore, timestamped phase annotations are provided to facilitate visualization during video playback.

To enable efficient processing and analysis, the dataset was preprocessed into shorter video clips. Each video was divided into non-overlapping clips of approximately 300 frames, corresponding to 12 seconds per clip at the original frame rate of 25 fps. This segmentation ensures manageable clip durations while preserving sufficient temporal context for downstream tasks. For each clip, phase and tool annotations were analyzed to determine dominant labels. Specifically, the most frequently occurring phase label within a clip was assigned as the clip's phase classification label, while tool presence was determined based on the maximum occurrence of each tool across frames. The resulting clips were saved in MP4 format using FFmpeg, maintaining the original resolution and frame rate. The preprocessing pipeline also generated metadata in JSON format, detailing clip paths, durations, and associated labels.

After preprocessing, the dataset consists of approximately 670 video clips, with a total duration of around 80 minutes. Each clip retains the original resolution and frame rate of 25 fps, ensuring consistency with the raw videos. The average clip length is 12 seconds, with slight variations due to the alignment of clip boundaries with video lengths. The resulting dataset is well-suited for training and evaluating models for surgical phase recognition and tool presence detection, offering a balanced distribution of surgical phases and tool usage patterns. This structured representation facilitates the development of deep learning models tailored to laparoscopic video analysis, contributing to advancements in computer-assisted surgery.

\noindent\textbf{Colonoscopic-web}
Colonoscopic-Web dataset contains 76 videos from colonoscopies documenting different types of gastrointestinal lesions. Each video is accompanied by annotated real data, including histopathologic findings, expert annotations, and calibration data from the recording system. Three categories, Adenoma, Hyperplasic and Serrated, were included in the dataset, representing the distribution of common polyp types. Long videos were cut into short 2-minute videos. The dataset is intended for use in studies of automated lesion detection and classification, with the goal of improving diagnostic accuracy and reducing the workload of clinicians by reducing the reliance on pigmented endoscopy in routine examinations.

\noindent\textbf{Endovis2019}
Endovis2019 dataset contains endoscopic video data from general surgical operating rooms, which were acquired during laparoscopic procedures at Heidelberg University Hospital and its affiliated hospitals. All surgeries were labeled frame-by-frame by surgical experts, and the labels include the phase of the surgery, surgical action, and instrument used. The recorded surgeries were mainly laparoscopic cholecystectomies. The dataset contains recorded videos of 30 different surgeries from at least three hospitals, and endoscopically captured videos of each procedure are provided. To ensure privacy, all additional extra-abdominal footage was masked by completely white frames. Tasks for this dataset include phase recognition, action classification, and tool classification. The phase recognition task entails classification based on the different stages of the surgical procedure. The dataset is labeled with a number of different stages ranging from the preparation stage to gallbladder traction. The action classification task involves categorizing surgical manipulation actions, including grasping, holding, cutting, and clipping actions. The instrument classification task required the classification of various tools used in surgery, covering a wide range of surgical instruments, like grasper, clipper, coagulation instruments, scissors, suction-irrigation, specimen bag and stapler. In addition, an “undefined instrument shaft” category was defined to label tools that are not explicitly categorized. Based on the label of each frame, the video is cut by frame classification, consecutive frames with the same label form a new video segment, and each new video segment is subsequently cut into 10-second videos.

\noindent\textbf{hyper-kvasir}
Hyper-Kvasir dataset is a gastrointestinal image and video dataset of 374 gastroenteroscopy and colonoscopy videos from Bærum Hospital, Norway, totaling 11.62 hours and more than 1 million frames. This dataset was categorized according to a hierarchical structure. First, the dataset was categorized with the Upper GI and Lower GI as the first tier. The data were then subdivided according to four main categories, which included anatomical landmarks, pathological findings, quality of mucosal views, and therapeutic interventions, which formed the second tier. These categories were further subdivided into 30 specific subcategories that belonged to the third tier. By combining these subcategories with labels for the upper and lower GI tract, each video is given a new label. And each long video was cut into small 5-second video clips based on its categorization and stored in the corresponding folder. In addition, the dataset contains 10,662 annotated images in JPEG format with images labeled in 23 different lesion types covering a wide range of normal and pathologic manifestations in different parts of the gastrointestinal tract. These images and videos provide a large amount of training data for the development of an AI-assisted gastrointestinal endoscopy analysis system, and can especially help researchers cope with the category imbalance problem that is common in medical data.

\noindent\textbf{kvasir-capsule}
Kvasir-Capsule dataset is an open dataset for gastrointestinal endoscopic video analysis. The dataset contains 4,741,621 pieces of data, including 47,238 labeled images with bounding boxes, 43 labeled videos, and 74 unlabeled videos. In addition, 4,694,266 unlabeled images can be extracted from all videos. The labeled images in the dataset are extracted from frames in the original long video. Based on the information of these labeled frames, the original video was cut into short videos every 5 seconds and kept consistent with the labels of the original long video. There were 47,238 labeled images in the dataset, which were classified into 14 different categories and divided into two main categories, Anatomy and Luminal findings. The Anatomy includes anatomical landmarks associated with the gastrointestinal tract, such as the pylorus, ileocecal valve, and ampulla of Vater, while the Luminal findings covers pathologic changes occurring in the gastrointestinal tract, such as Normal clean mucosa, reduced mucosal view, blood-fresh, blood-hematin, and so on. These images are widely used to study gastrointestinal diseases and help improve the diagnostic accuracy of endoscopy. At the same time, due to the imbalance in the number of images of various diseases in the dataset, especially for some of the rarer diseases, researchers need to employ effective machine learning methods that are able to learn from the limited training data, especially for the recognition of rare diseases.

\noindent\textbf{CholecT50}
CholecT50 is a dataset of endoscopic videos of laparoscopic cholecystectomy procedures for fine-grained action recognition, designed to facilitate research on action recognition techniques in laparoscopic surgery. The dataset contains 45 videos from the Cholec80 dataset and 5 videos from the internal dataset Cholec120, all of which are finely labeled with the triad of information (<instrument, verb, target>) for each surgical action as well as the corresponding stage labels. The dataset also provides the spatial annotation of the instrument tips (bounding boxes) in the 5 videos and the frame-triad matching labels in all videos. In addition, each frame in the videos was extracted at a frequency of 1 frame per second and annotated with detailed surgical movements, aiming to support the study of movement recognition algorithms in laparoscopic surgery.
The dataset was split into four tasks: phase classification, instrument classification, verb classification and target classification. The task division is based on the JSON annotation document for each frame, and each frame is labeled in detail, with several consecutive frames having the same label. To accommodate model training, all video clips are segmented into several short videos of 5 seconds with a frame rate of 10, and the labels are rearranged to map the labels of the four tasks to the range of 0-31  in order to uniformly standardize the training data.

\noindent\textbf{LDPolyVideo}
LDPolypVideo dataset is a large and diverse dataset of colonoscopic polyp detection. It contains colonoscopy videos collected from routine clinical colonoscopies with all patient-related metadata removed. The dataset contains 160 videos totaling 40,266 frames, of which 33,884 frames contain polyps. To increase the diversity of the dataset, the data includes not only clear images of polyps, but also covers motion blurring due to camera movement, bowel folding and bowel peristalsis. The dataset was divided into training, validation and test sets. The images in the validation and test sets were labeled to indicate whether each frame contained a polyp or not, with a label value of 0 or 1, respectively. For these frames, the video is synthesized into video clips with fps of 25 and cut into short 5-10 second segments based on the frame-level labels. The training set is then divided based on the labeling of the entire video, containing videos with polyps and videos without polyps. For videos with polyps, the video is split evenly into 2 to 4 segments and each segment is compressed into a 10-second video to ensure that polyps are contained within each segment. For videos without polyps, the videos were cut into short segments of 5 to 10 seconds. This treatment ensures that each segment is correctly labeled and improves the diversity of the dataset and the generalization ability of the model.

\noindent\textbf{JIGSAWS}
JIGSAWS dataset for modeling surgical activity in human motion was collected by Johns Hopkins University in collaboration with Intuitive Surgical. The dataset was collected using the da Vinci Surgical System from eight surgeons of varying skill levels, who performed five repetitions of three basic surgical tasks: suturing, knotting, and threading, which are standard components of most surgical skills training courses, on a tabletop model. The JIGSAWS dataset is comprised of three main components: the first component is kinematic data, which includes data that describes the operator's movement in terms of Cartesian positions, orientations, velocities, angular velocities and gripper angle describing the motion of the manipulators.; the second component is video data using stereoscopic video; and the third component is manually labeled data that includes gestures and skills. The dataset contains 15 labeled gestures in total, each corresponding to a specific task segment during surgery. The specific 15 labels include \textit{G1:}"grasping the needle with the right hand", \textit{G2:}"positioning the needle", \textit{G3:}"passing the needle through the tissue", \textit{G4:}"passing the needle from the left hand to right hand”, \textit{G5:}"Grasp the needle and move to the center", \textit{G6:}"Pull the suture with the left hand", \textit{G7:}"Pull the suture with the right hand", \textit{G8:}"Adjust the direction of the needle", \textit{G9:}"Use your right hand to help tighten the stitch", \textit{G10:}"Loosen more stitches", \textit{G11:}”Drop the stitch and move to the end", \textit{G12:}"Grab the needle with your left hand", \textit{G13:}"Make a C-shape around your right hand", \textit{G14:}"Grab the seam allowance with your right hand", \textit{G15:}"Pull the stitch with both hands". This labeling information was stored in a transcription file for each task, which consisted of threading the needle, sewing, and tying the knot.

\noindent\textbf{AlxSuture}
AIxSuture dataset was collected to analyze the effectiveness of training guided by virtual reality head-mounted displays. It contains 314 videos documenting students performing open surgical suturing in a simulated environment and categorizing them into proficient, novice, and intermediate categories based on skill scores. Skills were scored using the Objective Structured Assessment of Technical Skills scale, with eight skill categories, and the sum of the scores from all categories formed a global rating GRS ranging from 8 to 40. Mean values were obtained through a preliminary analysis of the pairwise Pearson correlation coefficients between the scores of the three assessors. For each video, the ratings of the three assessors were averaged and then categorized into three categories: novice, intermediate, and proficient. Due to the long duration of the videos, comprehension was required for long videos, which were divided equally into 2 to 4 equal parts and eventually compressed into 10-second segments for processing and analysis.

\noindent\textbf{AVOS}
Annotated Videos of Open Surgery (AVOS) collected 1997 open surgery videos from YouTube, expanding 23 surgical procedure types and 50 countries over the last 15 years. Among them, 326\footnote{30 video urls have expired, so we only have 296 annotated videos} videos are annotated with the action of the scene every five seconds. We first download the data from the dataset's homepage (\url{https://research.bidmc.org/surgical-informatics/avos}) and split the annotated videos to 5-second short clips, retaining the original frame rates and resolutions, with five action labels (cutting, suturing, typing, abstain and background). Finally, 27,745 clips with a total of 38.53 hours and 3,582,276 frames are obtained. The videos exhibit variable frame rates due to its multi-source nature, with the predominant FPS distribution concentrated within the range of 25 to 30.


\end{document}